\documentclass[11pt]{article}

\usepackage[final]{acl}
\usepackage{amsmath}
\usepackage{amssymb}
\usepackage{times}
\usepackage{latexsym}
\usepackage{multirow}
\usepackage{booktabs}
\usepackage{bm}
\usepackage[linesnumbered,ruled,vlined]{algorithm2e}
\usepackage{amsmath, bm, amssymb}
\usepackage[T1]{fontenc}

\usepackage[utf8]{inputenc}

\usepackage{microtype}

\usepackage{inconsolata}

\usepackage{graphicx}

\title{Punctuation-aware Hybrid Trainable Sparse Attention \\for Large Language Models}

\author{
 \textbf{Junxiang Qiu\textsuperscript{1}},
 \textbf{Shuo Wang\textsuperscript{1}\thanks{Shuo Wang and Qi Tian are the corresponding authors}},
 \textbf{Zhengsu Chen\textsuperscript{2}},
 \textbf{Hengheng Zhang\textsuperscript{2}},
 \textbf{Jinda Lu\textsuperscript{1}},
 \textbf{Changcheng Li\textsuperscript{1}},
 \textbf{Qi Tian\textsuperscript{2}\textsuperscript{*}}
\\  
 \textsuperscript{1}University of Science and Technology of China,
 \textsuperscript{2}Huawei Inc.
\\  
 {\small  
   \texttt{\{qiujx, lujd, lichangcheng\}@mail.ustc.edu.cn, 
   shuowang.edu@gmail.com,
   \{chenzhengsu2, zhanghengheng55, tian.qi1\}@huawei.com}
 }  
}

\begin{document}
\maketitle
\begin{abstract}
Attention serves as the fundamental mechanism for long-context modeling in large language models (LLMs), yet dense attention becomes structurally prohibitive for long sequences due to its quadratic complexity. Consequently, sparse attention has received increasing attention as a scalable alternative. However, existing sparse attention methods rely on coarse-grained semantic representations during block selection, which blur intra-block semantic boundaries and lead to the loss of critical information. To address this issue, we propose \textbf{P}unctuation-aware \textbf{H}ybrid \textbf{S}parse \textbf{A}ttention \textbf{(PHSA)}, a natively trainable sparse attention framework that leverages punctuation tokens as semantic boundary anchors. Specifically, (1) we design a dual-branch aggregation mechanism that fuses global semantic representations with punctuation-enhanced boundary features, preserving the core semantic structure while introducing almost no additional computational overhead; (2) we introduce an extreme-sparsity-adaptive training and inference strategy that stabilizes model behavior under very low token activation ratios; Extensive experiments on general benchmarks and long-context evaluations demonstrate that PHSA consistently outperforms dense attention and state-of-the-art sparse attention baselines, including InfLLM v2. Specifically, for the 0.6B-parameter model with 32k-token input sequences, PHSA can reduce the information loss by 10.8\% at a sparsity ratio of 97.3\%.

\end{abstract}

\section{Introduction}

\begin{figure}[t]
  \includegraphics[width=\linewidth]{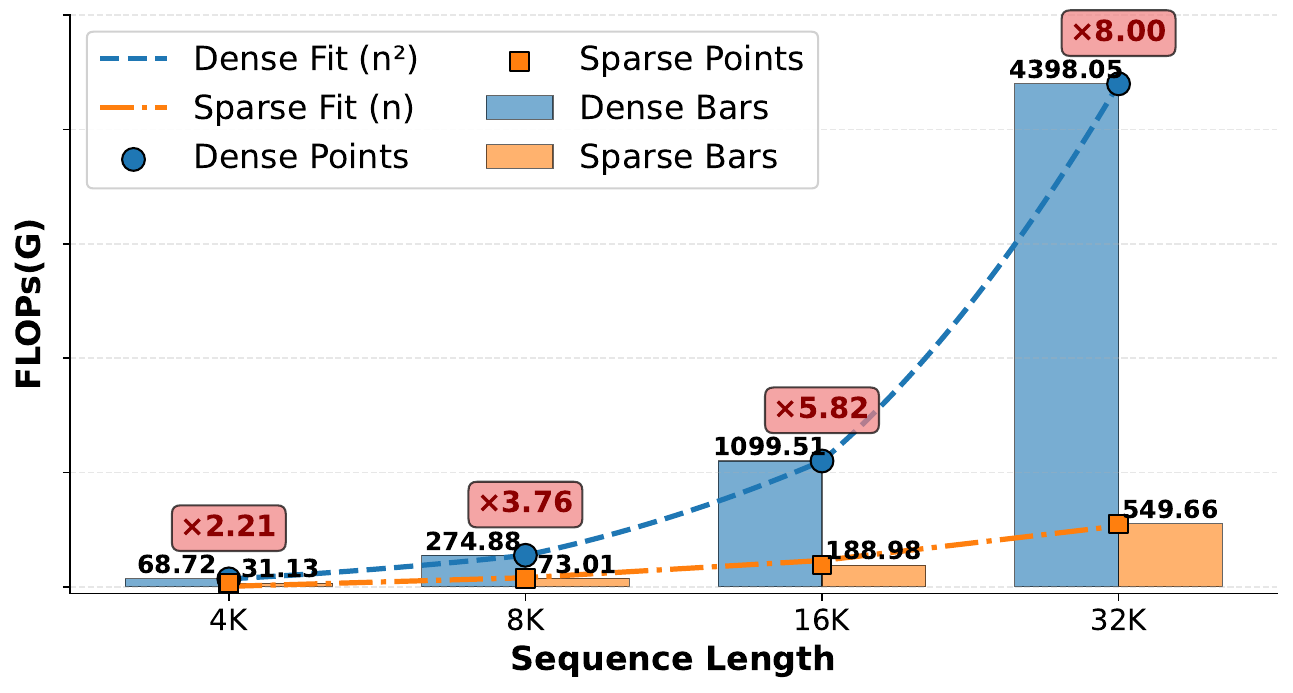}
  \caption{Comparison of the computational complexity of single-layer attention under dense (blue) and sparse (orange) settings for different sequence lengths.}
  \label{fig:flops}
\end{figure}

Long context processing \cite{bai2024longbench,li2024long} has emerged as a core capability enabler for large language models (LLMs) \cite{achiam2023gpt} to be deployed in real-world scenarios, with applications covering complex tasks such as code generation, agent interaction, and long-document understanding. 
At the architectural level, this capability is predominantly realized through the attention mechanism \cite{vaswani2017attention}, whose scalability directly determines the efficiency of long-context modeling. 

However, for a text sequence of length $L$, the dense attention mechanism requires each token to attend to all preceding tokens, resulting in a computational complexity of $O(L^2)$. As shown in the blue bars in Figure~\ref{fig:flops}, as $L$ increases, both inference latency and memory footprint grow rapidly, forming the primary hardware bottleneck for long-sequence processing. To overcome this limitation, recent works have focused on sparse attention mechanisms \cite{jiang2024minference,zhang2023h2o,yuan2025native,zhao2025infllm} that restrict attention computation to a subset of critical query-key pairs, reducing the overall complexity to $O(L)$. By avoiding redundant computations on irrelevant tokens, sparse attention significantly reduces computational and memory overheads, while also enhancing the model’s semantic focusing capability through native training. As illustrated in the orange bars in Figure~\ref{fig:flops}, sparse attention exhibits substantially better scalability with respect to sequence length, and its efficiency advantage becomes increasingly pronounced in long-context regimes. Meanwhile, the latest research indicates that sparse attention mechanisms with native training not only avoid inevitable degradation in model performance, but also have the potential to yield performance gains. For examples, DeepSeek’s NSA \cite{yuan2025native} and MiniCPM’s InfLLM v2 \cite{zhao2025infllm} employ sparse attention training-inference mechanism, and report strong performance on long-context benchmarks.

Despite the aforementioned progress, existing sparse attention methods still face two key limitations that restrict their practical applicability and limit further gains in model capability. First, during the key and value selection stage, existing methods aggregate consecutive tokens into a single representative vector through average pooling to reduce the computational and sorting overhead of query-key relevance estimation. This coarse-grained aggregation inevitably obscures fine-grained semantic distinctions within each block. Semantically critical elements such as key entities or logical hubs may be diluted by irrelevant tokens, leading to the loss of critical semantics and ultimately producing suboptimal selection results. Second, current sparse attention methods exhibit pronounced performance degradation under extremely low selection settings, which hinders the exploration of their full potential in the regime of extreme sparsity. For example, experiments on NIAH \cite{hsieh2024ruler} show severe information loss when the number of activated tokens is aggressively reduced. For instance, NSA recommends activating 3584 tokens (corresponding to approximately 11.2\% sparsity) in 32k long-sequence scenarios, while more aggressive sparsity regimes remain largely unexplored.

Motivated by the observation in SepLLM \cite{chen2024sepllm} that punctuation tokens (\emph{e.g.}, commas, periods, and semicolons) encode inherent semantic boundary information, we leverage their boundary-anchoring property to address the information loss caused by coarse-grained block aggregation. Punctuation tokens naturally divide text into logically coherent semantic segments and mark key semantic turning points across segments, providing a structural cue for preserving fine-grained semantics. Based on this insight, we design a dual-branch representative token aggregation mechanism for block-level token selection, in which a punctuation-aware mechanism is explicitly incorporated. This mechanism preserves critical semantic boundary information during block aggregation while avoiding the dilution of important features, thereby mitigating the semantic degradation introduced by average pooling. We refer to this punctuation-aware aggregation process as the core component of \textbf{P}unctuation-aware \textbf{H}ybrid \textbf{S}parse \textbf{A}ttention \textbf{(PHSA)}.

Meanwhile, beyond improving the accuracy of representative token selection, we further explore the feasibility of training and generation under extreme sparsity, which is a critical factor for achieving highly efficient sparse attention. Building upon the punctuation-aware sparse attention mechanism, we construct an extreme-sparsity-oriented training and inference framework that enables stable model behavior with a very small number of activated tokens. This framework substantially reduces computational and memory overhead on resource-constrained devices, breaks the efficiency limitations of existing methods, and expands the deployment scope of long-context processing. In summary, the contributions of our \textbf{PHSA} are threefold:

\begin{itemize}
\item We design a punctuation-aware hybrid aggregation mechanism, which adopts a dual-branch aggregation mechanism of "global semantic representation + punctuation semantic representation" to preserve critical semantic boundary information and improve the semantic accuracy of Top-K block selection.
\item We propose an adaptable training framework for extreme sparsity, systematically exploring lower activated token ratios, quantitatively analyzing performance boundaries, and providing theoretical support for lightweight deployment.
\item PHSA outperforms both dense attention and state-of-the-art baseline in general benchmarks and long-context benchmarks under both training and training-free scenarios.
\end{itemize}

\section{Related Work}
In this section, we introduce current sparse attention methods in terms of whether training is required.

\subsection{Training-free Sparse Attention} 
Training-free sparse attention methods aim to reduce inference latency by introducing sparsity into the attention mechanism without modifying the pre-trained dense attention backbone. These methods have made remarkable attempts in optimizing Key-Value (KV) cache usage and reducing computational overhead, but they still face significant challenges in translating theoretical advantages into practical efficiency improvements.
One major challenge is phase-limited sparsity. Methods such as H2O \cite{zhang2023h2o} apply sparsity during autoregressive decoding but require dense preprocessing during prefill. In contrast, methods like MInference \cite{jiang2024minference} focus on prefill sparsity. Neither of these approaches achieves acceleration across all inference phases, as the computational cost of at least one phase remains comparable to that of dense attention. This phase specialization impairs their acceleration capability in workloads dominated by prefill (\emph{e.g.}, book summarization and code completion) or decoding (\emph{e.g.}, long chain-of-thought \cite{wei2022chain} reasoning).

\subsection{Trainable Sparse Attention} 
Trainable sparse attention methods enable models to learn optimal sparse patterns during training, addressing performance degradation issues of post-hoc sparsity. But research on training-inference consistent sparse attention requires substantial computing resources, which leads to a slow pace of technological iteration. Representative works include ClusterKV \cite{liu2025clusterkv} with k-means clustering for token grouping and MagicPIG \cite{chen2024magicpig} using SimHash-based selection, though both contain untrainable discrete operations. HashAttention \cite{desai2024hashattention} adopts token-granularity selection but suffers from inefficient backpropagation. Recent advances focus on overcoming these limitations: NSA \cite{yuan2025native} designs dynamic hierarchical sparse strategies for end-to-end training; InfLLM v2 \cite{zhao2025infllm}, integrated into MiniCPM4 \cite{team2025minicpm4}, proposes a more streamlined fine-grained trainable sparse attention paradigm. These works strive to balance training efficiency, inference speed, and model performance, though the trade-off remains an open problem.

\begin{figure*}[t]
\centering
  \includegraphics[width=\linewidth]{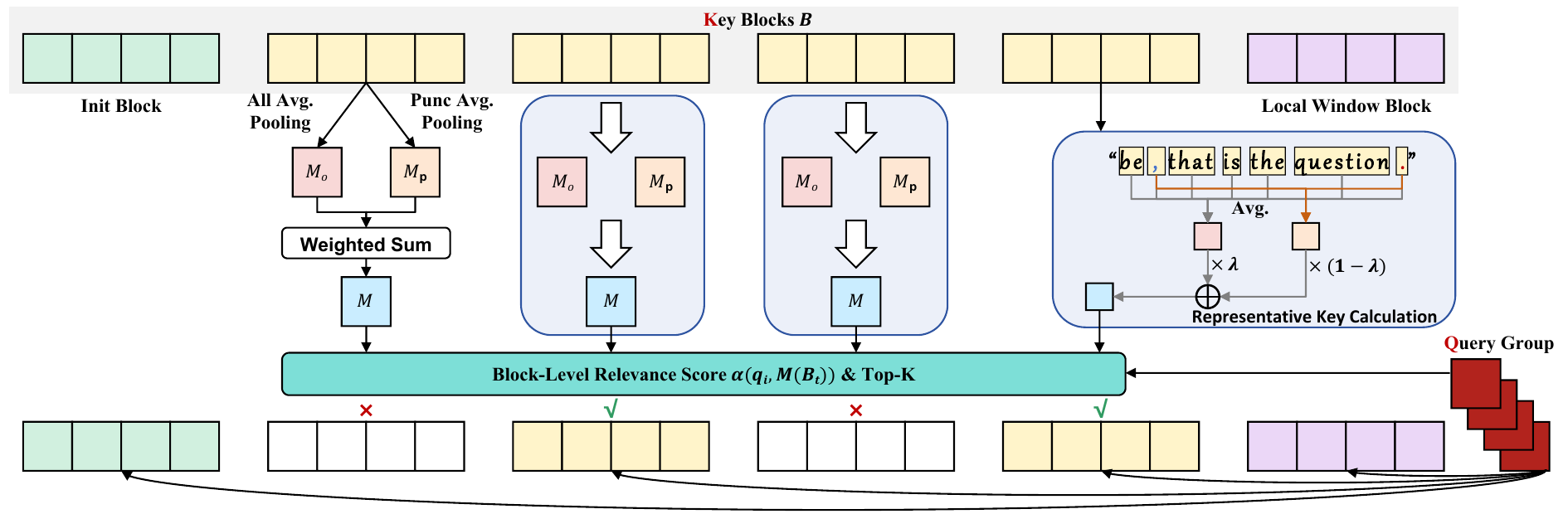}
  \caption{Pipeline of PHSA. By splitting key blocks, we first compute the dual-branch representative tokens for each block, then calculate the relevance score with the query. Finally, we select the Top-K blocks, which, together with the initial block and local window blocks, form the to-be-computed indices for sparse attention.}
  \label{fig:pipeline}
\end{figure*}

\section{Method}
This section elaborates on the structure of PHSA in detail. The subsequent subsections first introduce the preliminaries of our method, followed by a detailed exposition of the overall framework and core algorithms of PHSA.

\subsection{Preliminaries}
In each dense attention layer, the input sequence $ \bm{X} = \{ \bm{x}_1, \bm{x}_2, ..., \bm{x}_L \} $ is first mapped to query vectors $ \bm{Q} = \{ \bm{q}_1, \bm{q}_2, ..., \bm{q}_L \} \in \mathbb{R}^{L \times d_\text{k}} $, key vectors $ \bm{K} = \{ \bm{k}_1, \bm{k}_2, ..., \bm{k}_L \} \in \mathbb{R}^{L \times d_\text{k}} $, and value vectors $ \bm{V} = \{ \bm{v}_1, \bm{v}_2, ..., \bm{v}_L \} \in \mathbb{R}^{L \times d_\text{v}} $ through linear projection. 
The core idea of the attention mechanism in LLMs lies in:
each query token $\bm{q}_l \in \mathbb{R}^{d_\text{k}}$ (corresponding to the $l$-th position in the input sequence) computes semantic relevance scores with all key tokens from the preceding context $\bm{k}_{:l} = \{ \bm{k}_1, \bm{k}_2, ..., \bm{k}_l \in \mathbb{R}^{d_\text{k}}\}$, and then performs a weighted sum of the corresponding value tokens $\bm{v}_{:l} = \{ \bm{v}_1, \bm{v}_2, ..., \bm{v}_l \in \mathbb{R}^{d_\text{v}}\}$ based on these scores, ultimately yielding an output fused with contextual information.
Formally, for an input sequence of length $ l $, the attention output $\bm{o}_l$ at the $ l $-th position can be defined as:
\begin{equation}
  \label{eq:1}
  \bm{o}_l = \sum_{i=1}^{l} \bm{\alpha}(\bm{q}_l, \bm{k}_i) \cdot \bm{v}_i,
\end{equation}
in which
\begin{equation}
  \bm{\alpha}(\bm{q}_l, \bm{k}_i) =  \text{Softmax} ( {\bm{q}_l^\top \bm{k}_i}/{\sqrt{d_\text{k}}}  ),
\label{eq:2}
\end{equation}
where $\bm{\alpha}(\bm{q}_l, \bm{k}_i)$ denotes the attention weight between the query vector $\bm{q}_t$ and the $i$-th key vector $\bm{k}_i$, which measures the degree of semantic relevance between them; $\sqrt{d_\text{k}}$ is the dot-product normalization factor.

In long context processing scenarios, the computational complexity of the dense attention mechanism grows quadratically with the sequence length ($O(L^2)$, where $L$ denotes the sequence length), gradually becoming the core bottleneck limiting model inference efficiency. By only performing attention computations on a subset of contextual tokens, sparse attention provides a critical solution to break through this bottleneck.

\subsection{Punctuation-aware Hybrid Sparse Attention}

The pipeline of \textbf{P}unctuation-aware \textbf{H}ybrid \textbf{S}parse \textbf{A}ttention (\textbf{PHSA}) is shown in Figure~\ref{fig:pipeline}. 
First, we partition the key sequence into non-overlapping blocks of equal length. To preserve essential local context, the initial block and a local window around the query are assigned the highest priority and are always retained. The remaining intermediate blocks are treated as candidates for subsequent selection. For each candidate block, \textbf{PHSA} constructs a representative token by integrating two complementary semantic views: a global semantic representation and a punctuation-aware semantic representation. The global representation is obtained via average pooling over all tokens within the block, while the punctuation-aware representation is computed by pooling over punctuation tokens only. These two representations are then combined through a weighted aggregation to form a unified representative token for the block. Next, the relevance between the query and each representative token is evaluated, and a ranking-based strategy is employed to select the Top-K most relevant blocks. Attention is finally computed between the query and the keys and values within the selected blocks, producing the final attention output. In the following sections, we provide detailed descriptions of each component of \textbf{PHSA} and its implementation.

\noindent\textbf{Block Segmentation}: 
To optimize memory access efficiency, we adopt a block-level key-value caching strategy: 
Specifically, the key sequence $ \bm{K} $ is partitioned into non-overlapping blocks of equal size, each containing $ m $ consecutive tokens. This results in a set of key blocks $ \mathcal{B} = \{ \bm{B}_0, \bm{B}_1, \ldots, \bm{B}_{T-1} \} $,
where $ T =  \lfloor {L}/{m}  \rfloor $ and the $ t $-th key block $\bm{B}_t$ is defined as:
\begin{equation}
  \label{eq:3}
     \bm{B}_t = \bm{K}_{t \cdot m : (t+1) \cdot m}.
\end{equation}

The corresponding value tokens $ \bm{V} $ are accessed only after block selection and are therefore omitted here for clarity.

\noindent\textbf{Punctuation Token Selection}:
Punctuation marks (e.g., commas, periods, and semicolons) serve as semantic boundary markers in natural language, effectively dividing sentence- or phrase-level semantic units. To leverage this property, we first automatically extract punctuation-related tokens from the tokenizer vocabulary to construct a punctuation token set $\mathcal{P}$. Given the tokenizer vocabulary \( \mathcal{V} \), where each element corresponds to a token ID, the punctuation token set \( \mathcal{P} \) is formally defined as:
\begin{equation}
  \label{eq:4}
    \mathcal{P} = \{ p \in \mathcal{V} \mid \bm{P}(\bm{Z}(p)) = 1 \},
\end{equation}
where $\bm{Z}(p)$ denotes the set of character strings decoded from the token ID $p$. $\mathcal{P} \subseteq \mathcal{V}$ ensures that only punctuation tokens existing in the vocabulary are included. $\bm{P}(\cdot)$ is a punctuation judgment function used to filter out pure punctuation tokens:
\begin{equation}
\bm{P}(\bm{Z}(p)) = 
\begin{cases} 
1, & \bm{Z}(p) \in \Gamma, \\
0, & \bm{Z}(p) \notin \Gamma,
\end{cases}
\label{eq:5}
\end{equation}
where $\Gamma$ denotes the set of character strings for all punctuation marks. $\mathcal{P}$ provides fine-grained semantic boundary cues for block-level representation learning and relevance evaluation in the subsequent block selection stage.
To illustrate this process, we provide an example to illustrate these operations. Given a tokenized sentence: 
``\textit{To(1249) be(387) or(476) not(537) to(311) be(387) \textcolor{blue}{,(11)} that(429) is(374) the(279) question(3405) \textcolor{red}{.(13)}}''. 
In this case, tokens \textcolor{blue}{11} and \textcolor{red}{13} are identified as punctuation tokens in the punctuation token set \( \mathcal{P} \).

\noindent\textbf{Representative Tokens Calculation}:
To enable accurate and efficient block selection, we construct block-level semantic representations and compute their relevance to the query. The detailed formulation of each component is described as follows.

For each key block $\bm{B}_t$, we design two mean-pooling-based semantic representations to characterize its internal semantics from complementary perspectives: a global semantic representation and a punctuation semantic representation.

\noindent (1) Global Semantic Representation ($\bm{M}_0$).
The global representation captures the overall semantic content of the block and is defined as:
   \begin{equation}
     \label{eq:6}
     \bm{M}_0(\bm{B}_t) = \frac{1}{m} \sum_{i=t \cdot m}^{(t+1) \cdot m - 1} \bm{K}_i.
   \end{equation}

\noindent (2) Punctuation Semantic Representation ($\bm{M}_\text{p}$).
The punctuation-enhanced representation emphasizes semantic boundary information associated with punctuation tokens within the block and is defined as:
\begin{equation}
     \label{eq:7}
     \bm{M}_\text{p}(\bm{B}_t) = \frac{\sum_{i=t \cdot m}^{(t+1) \cdot m - 1} \bm{I}(\bm{x}_i \in \mathcal{P}) \cdot \bm{K}_i}{\sum_{i=t \cdot m}^{(t+1) \cdot m - 1} \bm{I}(\bm{x}_i \in \mathcal{P})},
\end{equation}
where $ \bm{I}(\cdot) $ is an indicator function (returns 1 if the condition is satisfied, otherwise 0); $ \bm{x}_i $ is the token ID of the $ i $-th token in the input sequence, and $ \bm{I}(\bm{x}_i \in \mathcal{P}) $ indicates whether the token is a punctuation token; $ \bm{K}_i $ is the $ i $-th row of the key vector matrix $ \bm{K} $. If the block does not contain any punctuation tokens, then $ \bm{M}_\text{p}(\bm{B}_t) = \bm{M}_0(\bm{B}_t)$.

The relevance score between the query token $ \bm{q}_i $ and the key block $ \bm{B}_t $ fuses the global semantic representation and the punctuation-enhanced representation through a gating mechanism, Eq.~\eqref{eq:2} can be rewritten in the following form:
\begin{equation}
  \begin{aligned}
  \bm{\alpha}(\bm{q}_i, \bm{M}(\bm{B}_{t})) = \text{Softmax}( \frac{\bm{q}_i^\top   ( {\bm{M}(\bm{B}_{t})}  )}{\sqrt{d_\text{k}}}), \\
  {\bm{M}(\bm{B}_{t})} = \lambda  \bm{M}_0(\bm{B}_{t}) + (1-\lambda) \bm{M}_\text{p}(\bm{B}_{t}),
  \end{aligned}
  \label{eq:8}
\end{equation}
where ${\bm{M}(\bm{B}_{t})}$ is the representative key of ${\bm{B}_{t}}$, and $ \lambda \in [0,1] $ is a gating parameter that balances the contributions of the two representations.

\noindent\textbf{Block Selection:}
Based on the block-level relevance score \( \bm{\alpha}(\bm{q}_i, \bm{M}(\bm{B}_t)) \), we select the Top-K key blocks with the highest scores. To ensure the preservation of essential contextual information, we introduce a priority mechanism in which the initial block and local window blocks are assigned the highest priority and are always retained in the attention set. This design guarantees that high-contribution blocks are consistently included in attention computation, mitigating the potential loss of critical information caused by sparse selection.

\noindent\textbf{Attention Calculation}:
The keys and values corresponding to the finally selected blocks are respectively combined into lightweight to-be-computed vectors $\bm{\hat{K}}$ and $\bm{\hat{V}}$ via concatenation. Meanwhile, the dense attention output as shown in Eq.~\eqref{eq:1} and Eq.~\eqref{eq:2} can be expressed in the following sparse attention output $\bm{\hat{o}}_l$ form:
\begin{equation}
  \label{eq:10}
  \bm{\hat{o}}_l = \sum_{i=1}^{l} \bm{\alpha}(\bm{q}_l, \bm{\hat{k}}_i) \cdot \bm{\hat{v}}_i,
\end{equation}
in which
\begin{equation}
  \bm{\alpha}(\bm{q}_l, \bm{\hat{k}}_i) =  \text{Softmax} ( {\bm{q}_l^\top \bm{\hat{k}}_i}/{\sqrt{d_\text{k}}}  ),
\label{eq:11}
\end{equation}
where $\bm{\hat{k}}_i$ and $\bm{\hat{v}}_i$ are selected for sparse attention calculation, and they belong to $\bm{\hat{K}}$ and $\bm{\hat{V}}$.

\begin{table*}[t]
\scriptsize
\centering
\begin{tabular}{cc|c|ccccc}
\toprule
\multirow{2}{*}{\textbf{Training Top-K}} & \multirow{2}{*}{\textbf{Training Method}} & \multirow{2}{*}{\textbf{Inference Method}} & \multicolumn{5}{c}{\textbf{Inference Top-K}}                                  \\  
                                        &                                           &                                            & \textbf{1}    & \textbf{2}    & \textbf{4}    & \textbf{8}    & \textbf{16}   \\ \midrule
\multicolumn{2}{c|}{\multirow{2}{*}{\textbf{N/A}}}             & \textbf{PHSA}                              & \textbf{83.5} & \textbf{96.4} & \textbf{98.4} & \textbf{99.8} & 99.8          \\
                                        &                                           & \textbf{InfLLM v2 \cite{zhao2025infllm}}                         & 80.6          & 94.0            & 97.0            & 99.4          & \textbf{100.0}           \\ \midrule
\multicolumn{1}{c|}{\textbf{0}}                              & \multicolumn{2}{c|}{\multirow{6}{*}{\textbf{PHSA}}}                                     & 0.0             & 0.0             & 0.0             & 0.0             & 0.0             \\
\multicolumn{1}{c|}{\textbf{1}}                              & \multicolumn{2}{c|}{}                                                                   & 79.6          & 91.4          & 97.6          & 99            & 99.4          \\
\multicolumn{1}{c|}{\textbf{2}}                              & \multicolumn{2}{c|}{}                                                                   & \textbf{81.8} & \textbf{95.0}   & \textbf{98.8} & \textbf{99.8} & 99.8          \\
\multicolumn{1}{c|}{\textbf{4}}                              & \multicolumn{2}{c|}{}                                                                   & 77.8          & 87.2          & 94.2          & 98.8          & 99.4          \\
\multicolumn{1}{c|}{\textbf{8}}                              & \multicolumn{2}{c|}{}                                                                   & 74.6          & 87.4          & 94.8          & 98.6          & 99.4          \\
\multicolumn{1}{c|}{\textbf{16}}                             & \multicolumn{2}{c|}{}                                                                   & 77.8          & 89.6          & 95.2          & 99.4          & \textbf{100.0}  \\ \midrule
\multicolumn{1}{c|}{\textbf{2}}                              & \multicolumn{2}{c|}{\textbf{InfLLM v2}}                                                 & 78.6          & 89.6          & 97.6          & 99.6          & 100.0           \\ \midrule
\multicolumn{2}{c|}{\multirow{2}{*}{\textbf{Dense}}}          & \textbf{PHSA}                              & \textbf{79.8} & 86.8          & \textbf{93.4} & \textbf{99.2} & \textbf{98.8} \\
                                        &                                           & \textbf{InfLLM v2}                         & 76.4          & \textbf{88.0}   & 93.6          & 98.4          & 98.4     \\ \bottomrule    
\end{tabular}
\caption{Comparison of NIAH scores across different combinations of training or inference methods (Dense, PHSA, and InfLLM v2) for a text length of 4k tokens. We focus on demonstrating the cross-matching effects between PHSA's training and generation under different Top-K values. And N/A denotes direct adoption of the Qwen3-0.6B-Base model.}
  \label{tab:lowtopk}
\end{table*}

\begin{table*}[t]
\scriptsize
\centering
\begin{tabular}{l|cc|cccccc}
\toprule
\textbf{Training Tokens}       & \multicolumn{2}{c|}{\textbf{N/A}} & \multicolumn{6}{c}{\textbf{20B}}             \\ \midrule
\textbf{Method}       & \multicolumn{2}{c|}{\textbf{Qwen-0.6B-Base}}     & \textbf{Dense}     & \textbf{Dense}     & \textbf{PHSA}     & \textbf{PHSA}     & \textbf{InfLLM v2} & \textbf{InfLLM v2}     \\ 
\textbf{Inference Top-K}       & \textbf{N/A}      & \textbf{2}     & \textbf{N/A}     & \textbf{2}     & \textbf{N/A}     & \textbf{2}     & \textbf{N/A} & \textbf{2}     \\ \midrule
\textbf{gsm8k}       & 56.18           & 49.05          & \fbox{54.97}          & 54.13          & \textbf{56.56} & 54.81          & 52.54      & 51.33          \\
\textbf{mathqa}      & 36.72           & 36.68          & 40.10           & 40.03          & 40.07          & 40.07          & \fbox{40.77}      & \textbf{40.94} \\
\textbf{math}        & 25.02           & 24.56          & \textbf{24.74} & 23.30           & \fbox{24.60}           & 19.12          & 21.80       & 21.80           \\
\textbf{arc\_c}      & 33.45           & 33.28          & 40.44          & 40.36          & \fbox{40.87}          & \textbf{41.13} & 39.33      & 39.51          \\
\textbf{arc\_c\_zh}  & 30.97           & 31.14          & \fbox{33.87}          & \textbf{33.96} & 33.70           & 33.36          & 33.53      & 33.36          \\
\textbf{arc\_e}      & 65.66           & 65.74          & \textbf{72.31} & \fbox{72.22}          & 72.20           & 71.97          & 71.89      & 71.51          \\
\textbf{c-valid}     & 54.90            & 54.61          & 55.87          & 56.02          & \textbf{57.80}  & \fbox{56.76}          & 56.69      & 55.79          \\
\textbf{cmmlu}       & 53.38           & 52.97          & \textbf{53.45} & 53.00          & 52.96          & \fbox{53.18}          & 52.61      & 52.93          \\
\textbf{hellaswag}   & 40.99           & 40.99          & 39.75          & 39.78          & \textbf{39.87} & \fbox{39.82}          & 39.72      & 39.64          \\
\textbf{humaneval}   & 28.05           & 28.05          & \textbf{37.80} & \fbox{35.98}          & 35.37          & 32.32          & 32.93      & 34.76          \\
\textbf{lambada}     & 53.56           & 53.79          & 50.16          & 50.09          & \textbf{50.30} & \fbox{50.22}          & 49.84      & 49.70          \\
\textbf{mmlu}        & 52.69           & 51.91          & \fbox{53.42}          & 53.09          & \textbf{53.43} & 53.17          & 52.95      & 52.68          \\
\textbf{pipa}        & 69.64           & 69.75          & 69.75          & 69.48          & \textbf{70.13} & 69.86          & \fbox{70.08}      & 70.02          \\
\textbf{xstorycloze} & 59.23           & 59.43          & 58.84          & \textbf{58.97} & 58.50           & 58.44          & 57.97          & \fbox{57.97}      \\
\textbf{bbh}         & 40.65           & 36.25          & \textbf{38.81} & 32.13          & \fbox{37.83}          & 34.71          & 38.07          & 34.39          \\   \midrule
\textbf{Avg.}        & 46.74           & 45.88          & \textbf{48.29} & 47.50          & \fbox{48.28}          & 47.26          & 47.38          & 47.09  \\ 
\bottomrule
\end{tabular}
\caption{Scores of Qwen3-0.6B-Base (initial training point) and differently trained models (20B training tokens) on general benchmarks. A Inference Top-K value of N/A indicates that dense attention is adopted during inference. Boldface and boxed values denote the best and second-best values.}
\label{tab:20b}
\end{table*}

\section{Experiments}
In this section, we evaluate the proposed PHSA method and perform a comparative analysis with dense attention and InfLLM v2. Our experiments cover both training-based and training-free paradigms, with comprehensive evaluations conducted on general benchmarks and long-context benchmarks. We aim to address the following research questions (\textbf{RQ}):

\noindent\textbf{RQ1}: Why prefer lower Top-K?

\noindent\textbf{RQ2}: Is PHSA effective on general benchmarks?

\noindent\textbf{RQ3}: Is PHSA effective on long-context benchmarks?

\noindent\textbf{RQ4}: What is the impact of punctuation marks across different languages?

\begin{table*}[t]
\scriptsize
\centering
\begin{tabular}{l|cccccccccc}
\toprule
\textbf{Training Tokens}       & \multicolumn{10}{c}{\textbf{100B}}                                                                                                                             \\ \midrule
\textbf{Method}       & \textbf{Dense}     & \textbf{Dense} & \textbf{PHSA}    & \textbf{PHSA} & \textbf{InfLLM v2}    & \textbf{InfLLM v2}    & \textbf{PHSA}    & \textbf{PHSA}    & \textbf{InfLLM v2}    & \textbf{InfLLM v2} \\ 
\textbf{Inference Top-K}       & \textbf{N/A}      & \textbf{2}     & \textbf{N/A}     & \textbf{2}     & \textbf{N/A}     & \textbf{2}     & \textbf{N/A} & \textbf{4}  & \textbf{N/A} & \textbf{4}     \\ \midrule
\textbf{gsm8k}       & 52.16          & 54.81       & 56.63          & 56.50        & 52.16          & 52.84          & 56.25          & 55.80           & \textbf{57.32} & \fbox{56.50}        \\
\textbf{mathqa}      & 41.91          & 42.24       & \textbf{42.91} & \fbox{42.58} & 41.81          & 42.31          & 42.48          & 42.38          & 42.24          & 41.98       \\
\textbf{math}        & 26.48          & 23.80        & 23.78          & 22.50        & 26.08          & 22.90           & \textbf{26.62} & \fbox{26.48} & 23.28          & 23.88       \\
\textbf{arc\_c}      & 41.30           & 41.13       & 41.13          & 41.04       & 41.97          & \textbf{42.15} & \fbox{41.97}   & 41.30           & 41.72          & 41.55       \\
\textbf{arc\_c\_zh}  & 34.30           & 34.22       & 33.02          & 32.68       & \textbf{35.67} & \fbox{35.41}    & 34.39          & 34.04          & 34.98          & 35.07       \\
\textbf{arc\_e}      & 71.84          & 71.97       & 71.13          & 71.21       & 72.52          & 72.31          & \fbox{72.64}   & \textbf{72.69} & 72.10           & 72.18       \\
\textbf{c-valid}     & 58.77          & 58.99       & \textbf{59.96} & \fbox{58.99} & 59.36          & 58.40           & 59.14          & 58.84          & 58.77          & 58.25       \\
\textbf{cmmlu}       & 55.38          & 55.44       & 55.16          & 55.16       & \textbf{55.90}  & \fbox{55.53}    & 55.23          & 54.96          & 55.48          & 55.51       \\
\textbf{hellaswag}   & \textbf{40.78} & \fbox{40.73} & 40.71          & 40.68       & 40.39          & 40.48          & 40.54          & 40.56          & 40.61          & 40.61       \\
\textbf{humaneval}   & 36.59          & 33.54       & 36.59          & 33.54       & 34.15          & 31.71          & \textbf{42.07} & \fbox{37.20}     & 36.59          & 35.37       \\
\textbf{lambada}     & 49.52          & 49.62       & \textbf{50.13} & \fbox{50.09} & 49.84          & 49.76          & 49.85          & 49.74          & 49.91          & 49.91       \\
\textbf{mmlu}        & \textbf{55.53} & \fbox{55.34} & 55.26          & 55.00          & 54.98          & 54.79          & 55.10           & 55.11          & 55.33          & 54.97       \\
\textbf{pipa}        & 70.95          & 70.84       & 70.62          & 70.62       & 71.06          & 70.89          & \textbf{71.27} & \fbox{71.22}   & 69.86          & 69.86       \\
\textbf{xstorycloze} & 58.90           & 58.90        & 58.90           & 58.97       & 59.50           & \fbox{59.56}    & 59.36          & \textbf{59.76} & 57.91          & 58.24       \\
\textbf{bbh}         & 40.02          & 33.99       & 39.56          & 33.37       & 39.78          & 26.26          & 32.11          & 32.78          & \textbf{40.55} & \fbox{40.02} \\  \midrule
\textbf{Avg.}        & 48.96          & 48.37       & 49.03          & 48.18       & 49.01          & 47.69          & \textbf{49.24}         & 48.81          & \fbox{49.11}          & 48.52      \\ \bottomrule
\end{tabular}
\caption{Scores of differently trained models (100B training tokens) on general benchmarks.}
  \label{tab:100b}
\end{table*}

\subsection{Experiment Settings}

\noindent\textbf{Datasets}.
The training dataset includes open-source datasets: dclm \cite{li2024datacomp}, map-cc \cite{du2024chinese}, ultrachat \cite{ding2023enhancing}, tuluv3 \cite{lambert2024tulu}, finemath \cite{liu2024finemath}, megamath \cite{zhou2025megamath}, as well as high-quality proprietary self-collected datasets related to education that are not publicly available.

\noindent\textbf{Model Configuration}.
We include both training and training-free experiments in the experimental section. Training is conducted on sequences with lengths of 4k and 32k tokens, which verifies the effectiveness of PHSA on general benchmarks and long-context benchmarks, respectively. For the 4k sequence length, our initialization block and local window block cover 16 and 128 tokens each. For the 32k sequence length, the corresponding values are 128 and 512 tokens. Each equally partitioned block consists of 16 tokens. Unless otherwise specified, our training is initialized from the Qwen3-0.6B-Base model, and the untrained sparse inference method is consistent with InfLLM v2. For the training-free experiments, we validate the performance of PHSA on two model scales: Qwen3-0.6B and Qwen3-8B \cite{yang2025qwen3}.

\noindent\textbf{Evaluation Metrics}.
To comprehensively verify the effectiveness of PHSA in general performance, long-context processing capability and information loss control, experiments were implemented on general benchmarks, LongBench \cite{bai2024longbench} and Needle-in-a-Haystack (NIAH) \cite{hsieh2024ruler} respectively. The general benchmarks cover 15 typical task categories, including mathematical reasoning (GSM8K \cite{cobbe2021training}, MathQA \cite{amini2019mathqa}, MATH \cite{hendrycks2021measuring}), commonsense question answering (ARC-C, ARC-C-ZH(Chinese translated version of ARC-C), ARC-E \cite{clark2018think}), comprehensive cognitive reasoning (MMLU \cite{hendrycks2020measuring}, CMMLU \cite{li2024cmmlu}, BBH \cite{suzgun2023challenging}), text coherence assessment (LAMBADA \cite{paperno2016lambada}, XStoryCloze \cite{lin2022few}), code generation (HumanEval \cite{chen2021evaluating}), and domain-specific tasks (C-Valid \cite{huang2023c}, HellaSwag \cite{zellers2019hellaswag}, PiPA \cite{kim2025pipa}), which span both English and Chinese bilingual scenarios. LongBench consists of four task modules, namely single-turn question answering (Single, including MQE, MQZ, NQA, QAS), few-shot learning (Fewshot, including LST, SSM, TRC, TQA), text summarization (Summarization, including GRP, MTN, QSM, VSM), and code-related tasks (Code, including LCC, RBP). NIAH primarily assesses the performance of sparse attention mechanisms in the preservation and retrieval of long-distance information, and it can most intuitively reflect the boundary of information loss under sparse attention.

\begin{table}[t]
\scriptsize
\centering
\begin{tabular}{c|c|ccc}
\toprule
\multirow{2}{*}{\textbf{Seq\_len}} & \multirow{2}{*}{\textbf{Method}} & \multicolumn{3}{c}{\textbf{Top-K}}           \\
                                   &                                  & 1             & 2             & 4           \\ \midrule
\multirow{2}{*}{\textbf{16k}}      & \textbf{InfLLM v2 \cite{zhao2025infllm}}               & 62.0            & 87.0            & 97.8        \\
                                   & \textbf{PHSA}                    & \textbf{69.0}   & \textbf{89.2} & \textbf{98.6}\\ \midrule
\multirow{2}{*}{\textbf{32k}}      & \textbf{InfLLM v2}               & 64.6          & 85.8          & 98.0          \\
                                   & \textbf{PHSA}                    & \textbf{68.0}   & \textbf{88.8} & \textbf{99.0} \\ \bottomrule
\end{tabular}
\caption{Comparison of NIAH scores of PHSA and InfLLM v2 inference methods for Qwen3-8B.}
\label{tab:8B}
\end{table}

\begin{figure*}[t]
  \includegraphics[width=\linewidth]{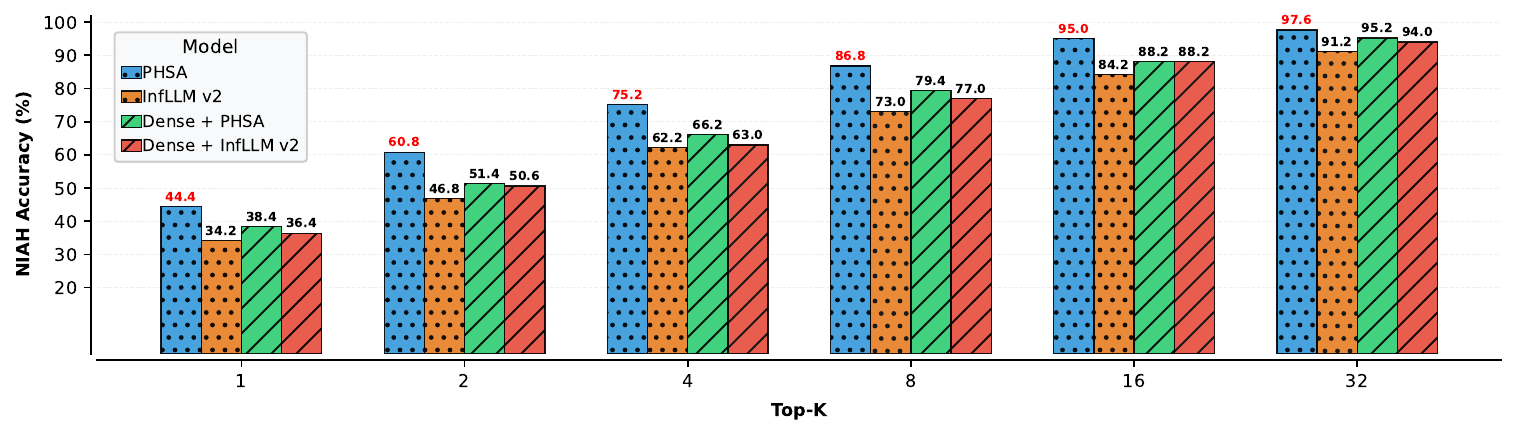}
  \caption{Comparison of NIAH scores for PHSA and InfLLM v2 (Training vs. Training-free) at 32k sequence.}
  \label{fig:long}
\end{figure*}

\begin{table*}[t]
\scriptsize
\begin{tabular}{l|cccc|cccc|cccc|cc|c}
\toprule
\multirow{2}{*}{\textbf{Method}} & \multicolumn{4}{c|}{\textbf{Single}}                      & \multicolumn{4}{c|}{\textbf{Fewshot}}                  & \multicolumn{4}{c|}{\textbf{Summarization}}               & \multicolumn{2}{c|}{\textbf{Code}} & \multirow{2}{*}{\textbf{Avg.}} \\ 
                                & \textbf{MQE}   & \textbf{MQZ}   & \textbf{NQA}   & \textbf{QAS}   & \textbf{LST} & \textbf{SSM}   & \textbf{TRC}  & \textbf{TQA}   & \textbf{GRP}   & \textbf{MTN}   & \textbf{QSM}   & \textbf{VSM}   & \textbf{LCC}         & \textbf{RBP}        &                                \\ \midrule
\textbf{Dense}                  & 34.53 & 22.13 & 7.16           & 20.47          & 25.67        & \textbf{28.12} & \textbf{58.5} & 74.90           & 13.11          & \textbf{15.37} & 15.33          & 12.23          & 30.47                & 32.05               & 27.86                          \\
\textbf{InfLLM v2}              & 31.02          & 21.85          & 10.18 & 20.98          & 28.00  & 26.36          & 52.0            & 77.71 & 13.19 & 15.30           & 15.54          & 12.65          & 26.98                & 31.80                & 27.40                          \\
\textbf{PHSA}                   & 33.95          & 20.63          & 8.11           & \textbf{22.46} & 26.75        & 26.70           & 55.5          & 76.94          & 13.17          & 15.06          & \textbf{15.55} & 12.75 & \textbf{31.08}       & \textbf{35.14}      & 28.13 \\ 
\textbf{PHSA\_en+zh}                   & \textbf{35.96}          & \textbf{22.22}          & \textbf{11.04}           & 20.67 & \textbf{30.50}        & 26.02           & 52.5          & \textbf{79.85}          & \textbf{13.45}          & 15.07          & 15.54 & \textbf{14.17} & 25.95       & 31.20      & \textbf{28.15} \\
\bottomrule
\end{tabular}
\caption{LongBench scores for models trained with Dense, InfLLM v2, PHSA and PHSA\_en+zh.}
  \label{tab:longbench}
\end{table*}

\subsection{Effect of PHSA}

\textbf{Low Top-K Preference (RQ1)}.
A large number of attention computations are performed on irrelevant tokens during model training. Sparse attention enables the model to filter out the most relevant tokens in the training process, thereby accelerating training and facilitating faster convergence simultaneously. However, an excessively low Top-K value often prevents the model from capturing all relevant tokens, which consequently leads to model performance collapse.
As shown in Table~\ref{tab:lowtopk}, in the scenario with training-inference consistent, training with a relatively low Top-K can reduce information loss in inference with a low Top-K, while an excessively small Training Top-K will degrade the model performance. Specifically, Training Top-K=2 serves as the optimal configuration, yielding an NIAH score of 95 at inference Top-K=2, which is significantly higher than the score of 91.4 when Training Top-K=1, and the performance drops noticeably when Training Top-K$\ge$4. Under the same training Top-K setting, our method (PHSA) achieves better performance; for instance, at Training Top-K=2, our method obtains a score of 81.8 at inference Top-K=1, outperforming InfLLM v2 with a score of 78.6. Thus, Training Top-K=2 or 4 is the locally optimal sparsity level for short-text training. In addition, PHSA also achieves higher accuracy than InfLLM v2, both on dense-trained models and when directly applied to the Qwen3-0.6B-Base model.

\textbf{Effect on general benchmarks (RQ2)}.
General benchmarks serve as an effective means to evaluate the fundamental capabilities of models. We present the dense inference scores and scores under training-inference consistent settings of PHSA, InfLLM v2 and dense attention with 20B and 100B training tokens.
Table~\ref{tab:20b} shows that PHSA outperforms InfLLM v2 (47.38) with an average score of 48.28 under both dense inference and training-inference consistent (Top-K=2) settings at 20B tokens, achieving performance comparable to dense attention (48.29).As shown in Table~\ref{tab:100b}, When trained with 100B tokens, PHSA still maintains superiority over InfLLM v2 under Top-K=2 and Top-K=4, and further achieves a leading average score of 49.24 at Top-K=4, which even surpasses the performance of dense attention (48.96).
Experiments in RQ2 indicate that compared with dense attention, PHSA has the potential to enhance model general capabilities under low Top-K setting.

\textbf{Effect on long-context benchmarks (RQ3)}.
Long-context benchmarks are important metrics for evaluating the performance of sparse attention methods. This is because sparse attention achieves greater efficiency gains in long-context scenarios, while also entailing a higher probability of information loss.
Table~\ref{tab:8B} demonstrates the consistent gains of PHSA over InfLLM v2 on NIAH for the 8B model, and also validates the effectiveness of the training-free paradigm in long-context scenarios.
We also conduct extensive training with Top-K set to 16. As evidenced by the NIAH metrics illustrated in Figure~\ref{fig:long}, training can further amplify the gains of PHSA. Specifically, PHSA reduces the loss of critical information by approximately 10\% compared with InfLLM v2.

Table~\ref{tab:longbench} reports the performance of dense attention, InfLLM v2 and PHSA on the LongBench benchmark. PHSA achieves competitive and superior performance over the two baselines on most subtasks, with an overall average score of 28.13, which outperforms dense attention (27.86) and InfLLM v2 (27.40). These results verify that PHSA can effectively boost the model's overall long-context modeling performance on LongBench.

\subsection{Additional Discussion}

\textbf{Cross-Lingual Impact (RQ4)}.
We note that PHSA delivers suboptimal performance on two Chinese-related metrics, namely arc\_c\_zh in Table~\ref{tab:100b} and MQZ in Table~\ref{tab:longbench}. We attribute this deficiency to the exclusive use of English punctuation marks during the training phase. To verify this conjecture and further demonstrate the effectiveness of PHSA across different languages, we incorporate Chinese punctuation marks for training and obtain the PHSA\_en+zh model presented in Table~\ref{tab:longbench}. It can be seen that the MQZ metric is improved from 20.63 to 22.2, which confirms that the introduction of additional punctuation indeed contributes to enhancing the model's performance on tasks for the corresponding language.

\section{Conclusion}
In this paper, we propose a Punctuation-aware Hybrid Sparse Attention (PHSA) mechanism, which takes punctuation tokens as natural semantic boundary anchors and designs a dual-branch aggregation mechanism integrating global semantic representations and punctuation-enhanced boundary features, effectively preserving core semantic information without additional computational overhead.
Extensive experimental results validate that the proposed PHSA outperforms dense attention and the state-of-the-art baseline InfLLM v2 across both general benchmarks (such as GSM8K, MMLU) and long-context evaluation tasks (including NIAH, LongBench) under both training and training-free paradigms.
PHSA not only provides a new effective solution for improving the efficiency and performance of long-context processing in large language models but also lays a theoretical and experimental foundation for the lightweight deployment of large language models on resource-constrained devices, thereby expanding the practical application scenarios of long-context processing.

\section*{Limitations}
Although PHSA can effectively reduce the frequency of information loss in sparse attention, we argue that the wholesale introduction of punctuation marks would introduce considerable noise. We believe that a screened subset of punctuation marks has the potential to raise the upper bound of PHSA performance; however, due to the constraints of paper length and computational resources, we will conduct an in-depth investigation into this direction in future work.

\bibliography{custom}

\newpage
\appendix
\onecolumn
\section*{Appendix}
\section{Algorithm of PHSA}
\label{sec:appendix}

The pseudocode for the attention component of PHSA during training is presented in Algorithm \ref{alg:phsa}.

\begin{algorithm}[h]
    \caption{Punctuation-aware Hybrid Sparse Attention (PHSA)}
    \label{alg:phsa}
    \SetKwInOut{Input}{Input}
    \SetKwInOut{Output}{Output}
    \SetKwFunction{Softmax}{Softmax}
    \SetKwFunction{TopK}{TopK}
    \SetKwFunction{Unique}{Unique}
    \SetKwFunction{IndexSelect}{IndexSelect}
    
    \Input{
        Query $\bm{Q}$, Key $\bm{K} \in \mathbb{R}^{L \times d_\text{k}}$, Value $\bm{V} \in \mathbb{R}^{L \times d_\text{v}}$, 
        Input Sequence $\bm{X}$, 
        Punctuation Set $\mathcal{P}$, 
        Block size $m$, 
        Top-K count $k$, 
        Window size $w$ (Local window), 
        Init count $n_\text{init}$ (Init block), 
        Gate param $\lambda$
    }
    \Output{Sparse Attention Output $\hat{\bm{O}}$}

    \tcp{1. Block Segmentation}
    Divide $\bm{K}$ into blocks $\mathcal{B} = \{ \bm{B}_0, \bm{B}_1, \dots, \bm{B}_{T-1} \}$, where $T = \lfloor L/m \rfloor$\;
    \tcp{2. Representative Tokens Calculation}
    Initialize block representations $\mathcal{M} = \emptyset$\;
    \For{$t  \leftarrow 0$ \KwTo $T-1$}{
        Calculate Global Rep: $\bm{M}_0(\bm{B}_t) = \frac{1}{m} \sum_{l \in \text{indices}(\bm{B}_t)} \bm{K}_l$ \;
        Identify punctuation: $\text{mask}_p = \{ \bm{I}(\bm{x}_l \in \mathcal{P}) \mid l \in \text{indices}(\bm{B}_t) \}$ \;
        \eIf{$\sum \text{mask}_p > 0$}{
            Calculate Punctuation Rep $\bm{M}_\text{p}(\bm{B}_t)$\;
            $\bm{M}(\bm{B}_t) = \lambda \cdot \bm{M}_0(\bm{B}_t) + (1-\lambda) \cdot \bm{M}_\text{p}(\bm{B}_t)$ \;
        }{
            $\bm{M}(\bm{B}_t) = \bm{M}_0(\bm{B}_t)$\;
        }
        $\mathcal{M}  \leftarrow \mathcal{M} \cup \{ \bm{M}(\bm{B}_t) \}$\;
    }
    Stack $\mathcal{M}$ to form compressed keys $\bm{K}_\text{comp} \in \mathbb{R}^{T \times d_\text{k}}$\;

    \tcp{3. Relevance Scoring }
    Compute scores $\bm{S} \in \mathbb{R}^{L \times T}$ where $\bm{S}_{i,t} = \bm{q}_i^\top \bm{M}(\bm{B}_t) / \sqrt{d_\text{k}}$\;

    Initialize selected block indices $\mathcal{I}_\text{total} = \emptyset$\;
    \ForEach{query token $\bm{q}_i$ (or parallelized)}{
        $\mathcal{I}_\text{init}  \leftarrow \{0, \dots, n_\text{init}-1\}$ 
        $\mathcal{I}_\text{local}  \leftarrow \{ \max(0, \lfloor i/m \rfloor - w/m), \dots, \lfloor i/m \rfloor \}$ 
        
        Apply causal mask to $\bm{S}_{i,:}$ (mask future blocks)\;
        Mask $\mathcal{I}_\text{init}$ and $\mathcal{I}_\text{local}$ positions in $\bm{S}_{i,:}$ with $-\infty$\;
        $\mathcal{I}_\text{Top-K}  \leftarrow \text{indices of Top-}k \text{ values in } \bm{S}_{i,:}$ 
        
        $\mathcal{I}_\text{final}  \leftarrow \Unique(\mathcal{I}_\text{init} \cup \mathcal{I}_\text{local} \cup \mathcal{I}_\text{Top-K})$\;
        Store $\mathcal{I}_\text{final}$ for reconstruction\;
    }

    \tcp{4. Attention Calculation}
    Map block indices $\mathcal{I}_\text{final}$ to token indices $\mathcal{I}_\text{tok}$\;
    Gather sparse KV: $\hat{\bm{K}}  \leftarrow \bm{K}[\mathcal{I}_\text{tok}], \hat{\bm{V}}  \leftarrow \bm{V}[\mathcal{I}_\text{tok}]$\;
    Compute output $\hat{\bm{o}}_i = \text{Softmax}(\bm{q}_i \hat{\bm{K}}^\top / \sqrt{d_\text{k}}) \hat{\bm{V}}$ \;
    
    \Return{$\hat{\bm{O}} = \{ \hat{\bm{o}}_1, \dots, \hat{\bm{o}}_L \}$}
\end{algorithm}

\end{document}